\documentclass[conference, letterpaper]{IEEEtran}
\usepackage{cite}
\usepackage{amsmath,amssymb,amsfonts}
\usepackage{algorithm,algorithmic}
\usepackage{graphicx}
\usepackage{textcomp}
\usepackage{xcolor}

\ifCLASSOPTIONcompsoc
\usepackage[caption=false,font=normalsize,labelfont=sf,textfont=sf]{subfig}
\else
\usepackage[caption=false,font=footnotesize]{subfig}
\fi

\def\BibTeX{{\rm B\kern-.05em{\sc i\kern-.025em b}\kern-.08em
    T\kern-.1667em\lower.7ex\hbox{E}\kern-.125emX}}
    
\usepackage{tikz}
\newcommand\copyrighttext{%
  \footnotesize \copyright 2021 IEEE. Personal use of this material is permitted. Permission from IEEE must be obtained for all other uses, in any current or future media, including reprinting/republishing this material for advertising or promotional purposes, creating new collective works, for resale or redistribution to servers or lists, or reuse of any copyrighted component of this work in other works.}
\newcommand\copyrightnotice{%
\begin{tikzpicture}[remember picture,overlay]
\node[anchor=south,yshift=10pt] at (current page.south) {\fbox{\parbox{\dimexpr\textwidth-\fboxsep-\fboxrule\relax}{\copyrighttext}}};
\end{tikzpicture}%
}

\begin{document}

\title{Discovering an Aid Policy to Minimize Student Evasion Using Offline Reinforcement Learning\\
}

\author{\IEEEauthorblockN{Leandro M. de Lima}
\IEEEauthorblockA{\textit{Graduate Program in Computer Science, PPGI
} \\
\textit{Federal University of Espirito Santo, UFES}\\
Vitória, Brazil \\
leandro.m.lima@ufes.br}
\and
\IEEEauthorblockN{Renato A. Krohling}
\textit{Graduate Program in Computer Science, PPGI}\\
\IEEEauthorblockA{\textit{Production Engineering Department, LABCIN} \\
\textit{Federal University of Espirito Santo, UFES}\\
Vitória, Brazil \\
rkrohling@inf.ufes.br}
}

\maketitle

\copyrightnotice

\begin{abstract}
High dropout rates in tertiary education expose a lack of efficiency that causes frustration of expectations and financial waste. Predicting students at risk is not enough to avoid student dropout. Usually, an appropriate aid action must be discovered and applied in the proper time for each student. To tackle this sequential decision-making problem, we propose a decision support method to the selection of aid actions for students using offline reinforcement learning to support decision-makers effectively avoid student dropout. Additionally, a discretization of student's state space applying two different clustering methods is evaluated. Our experiments using logged data of real students shows, through off-policy evaluation, that the method should achieve roughly 1.0 to 1.5 times as much cumulative reward as the logged policy. So, it is feasible to help decision-makers apply appropriate aid actions and, possibly, reduce student dropout.
\end{abstract}


\section{Introduction}
Globally the gross enrollment ratio in tertiary education increased from $19\%$ in 2000 to $38\%$ in 2018 according to Unesco \cite{Unesco}, but the expansion of the higher education system may not result in an increase in graduated professionals. In 2017, on average $33\%$ of the students fail to successfully complete the programs they undertake \cite{OECD2019education}. Especially in developing countries that have fragile economies, as in Latin America, effectiveness in transforming tertiary enrollment rates increase into the supply of high-skilled workers is key for their development. Evasion in tertiary education rates are important indicators of that efficiency and their results must be improved.

Student evasion incurs wasted resources, frustrations of expectations, and loss of personal, professional, and social potential. In public education institutions, it also represents an onus for society, especially for the financial waste it entails. Institutional policies and actions are extremely important to prevent that outcome. Identifying which students are at risk of evasion and what actions to take for each of them to minimize it is a complex problem that affects educational institutions. 
There are several reasons for students to abandon their undergraduate studies and those reasons, for the most cases, remain undetected by educational institutions until the moment a student initiates a transfer request, leave of absence request, or drops out. Institutions that can identify students with high evasion risk, and manage to successfully overcome student complains early on, may increase their graduation rates and contribute, ultimately, to the progress of society as a whole \cite{Balaniuk2011}.

\subsection{Related Works}

Researches in the educational data mining field mainly focus on classification, clustering, and visual data mining\cite{Aldowah2019} techniques to solve educational issues, as predicting students at risk. Most of the previous work in the student evasion field focuses on predictive tasks \cite{Sales2016, Kemper2020studentdropout, Alban2019dropoutreview}. Those approaches are an important step towards effective monitoring and help those students, but simply predicting who is at risk is not enough to solve the problem and minimize evasion. For such cases, it is necessary to identify a policy that defines the sequence of effective aid actions for each of the students at risk.  However, adequate aid actions at the right moment still mostly depend on the specialist’s correct decision due to a lack of appropriate decision support tools. Our approach is similar to \cite{ju2020edmofflineRL}, but their work is focused in identify critical decisions in interactive learning environments proposing a reinforcement learning based framework.

Many deep reinforcement learning (RL) algorithms have shown great results in sequential decision-making problems and this work is based on those methods to help decision-makers minimize student evasion. Deep reinforcement learning algorithms, as Deep Q-Network (DQN)\cite{mnih2015human}, Advantage Actor-Critic (A2C)\cite{mnih2015human}, Trust Region Policy Optimization (TRPO)\cite{Schulman2015}, Stochastic Lower Bounds Optimization (SLBO) \cite{Luo2019}, have shown advantages in learning human behaviors and strategies in several areas, providing good results in problems with high-dimensional action and state space. Usually, (online) reinforcement learning typically is employed in problems that can directly apply actions and observe the environment to acquire more data. However, data from previous interactions could have been already collected. Besides that, in some situations, there are cost, safety, or ethical issues in the usage of artificial intelligence (AI) decisions without a known AI quality to gather more data, as in health or educational field. Offline (or batched) reinforcement learning \cite{Agarwal2019, Fujimoto} seeks to overcome these limitations in exploration. 

Off-policy reinforcement learning methods, as DQN\cite{mnih2015human}, use data generated by an exploratory policy to learn an optimal policy \cite{sutton2018reinforcement}. Usually, those methods use an online sample collection to mitigate distributional shift issues, despite that, off-policy RL can learn reasonably well in offline problems\cite{Levine}. 

As the suggested actions often cannot be evaluated in the environment in offline problems, off-policy policy evaluation (OPE) can evaluate newly learned policies without interacting with the environment. OPE is an active field of research and multiples estimators can be applied, e.g. Sequential Weighted Doubly-Robust (SWDR) or Model Guided Importance Sampling Combining (MAGIC) \cite{Thomas}.

This work is also related to imitation learning and inverse reinforcement learning as they also try to learn how to mimic other policies. In imitation, learning the goal consists of finding a policy that directly mimics the observed policy, but it also suffers from distributional shift\cite{Reddy}. Inverse reinforcement learning tries to find a reward function that explains the policy behavior that generated the data\cite{Luceri}. However, both are inappropriate approaches as they require either access to an oracle, further online data collection, or an explicit distinction between expert and non-expert data\cite{Fujimoto}. Distributional shift also affects the model-based algorithms that suffer from it in the same way as the value function, since out-of-distribution state-action tuples can result in inaccurate prediction \cite{Levine}.

The proposed decision support method can not act directly in the environment. So, its recommendations can be reviewed or changed by the decision-maker before they are applied. Similar methodologies were applied to sepsis treatment \cite{Komorowski2018, Raghu2017}. Offline reinforcement learning can also be applied in many other sequential decision-making problems as in healthcare, spoken dialogue systems, self-driving cars, and robotics\cite{Levine}. 

 The goal of this work is to propose a decision support method for selection of aid actions for students based on a deep reinforcement learning approach aiming to help decision-makers of an educational institution. A method that recommends actions to aid students at risk of evasion to reduce the institutional evasion rate. Also, a novel case study is presented to illustrate the offline RL approach to automatically suggesting actions to aid those students with an analysis comparing the performance of the application using different evaluation methods and comparing the impact of different clustering methods on application performance.
 
 The remainder of this paper is organized as follows. In Section~\ref{sec:problem_statement}, key concepts of student dropout and offline RL are defined. Additionally, we provide an explanation of how to model the problem as a Markov decision process (MDP) and define a discrete state space for student dropout data. In Section~\ref{sec:methodology}, we propose a method to properly tackle the aforementioned problem. Next, the experimental results are presented in Section~\ref{sec:experimental_results}. Finally, Section~\ref{sec:conclusion} summarizes key findings of the experimental results and their implications. It also presents new directions for further investigation and future work.  

\section{Problem Statement}
\label{sec:problem_statement}

Student dropout or evasion are referenced here as any student that cannot complete their initiated studies and will not be awarded an academic degree. Our approach is based on a program perspective and any reason that caused a student to not graduating in a program is considered as dropout, e.g. transfer, withdrawal, poor performance.

This research aims to reduce the dropout rate by providing the most appropriate help for each student, according to their profile. Also, it is intended that the decision-maker has a reduction in the choice's workload, complexity, and pressure of who needs help, what action to take and when to apply that action to minimize student evasion.

\subsection{Markov decision process}
Time-varying state space can be described as a Markov decision process (MDP). In student evasion problem, it is not possible to directly observe the underlying state, therefore, that problem can be more precisely defined as a partially observable Markov decision process (POMDP). Due to simplification, we consider the state observation as a full representation of the state, so the student evasion problem is defined here as a fully-observed MDP. 

A MDP\cite{puterman1990markov} is characterized by $M = \{S,A,P,R\}$:
\begin{itemize}
    \item \textbf{State space} $S$. At each step $t$ (e.g. academic term), $s_t \in S$ defines a student state.
    
    \item \textbf{Action space} $A$. At each step $t$, an agent takes action $a_t \in A$ to modify the environment. In the student evasion problem, it consists of all combinations of the existence of a study plan and the types of student aid.
    
    \item \textbf{Transition function} $P(s_{t+1}|s_t, a_t)$. The probability $P : S \times A \times S \to \mathbb{R}$ of seeing state $s_{t+1}$ after taking action $a_t$ at state $s_t$.
    
    \item \textbf{Reward function} $R(s_t, a_t) \in \mathbb{R}$. A function $R: S \times A \to \mathbb{R}$ that returns the observed immediate feedback $r_t$ after transition $(s_t,a_t)$ at time $t$. Transitions to desired states generate positive rewards, transitions to undesired states negative ones.

\end{itemize}

In student evasion context, each student is represented as a trajectory $H = \{(s_1,a_1, r_1), (s_2,a_2, r_2), \dots, (s_T,a_T, r_T)\}$, that is a sequence  of state, action and reward tuples $(s_t,a_t, r_t)$  over $T$ academic terms. The agent seeks maximize the expected discounted return $R_t = \sum^{T}_{\tau = t} \gamma^{\tau-t}r_{\tau}$, where $\gamma \in [0,1]$ is a discount factor that trades-off the importance of immediate and future rewards.

\subsection{Offline (batch) reinforcement learning}

The goal in general reinforcement learning is to learn an optimal policy $\pi$ that observes the current state $s$ and selects an action $a$ which maximizes the sum of expected rewards. Yet, offline reinforcement learning can be defined as the task of learning the best possible policy from a fixed set $\mathcal{D}$ of \textit{a priori}-known transition samples or as a data-driven reinforcement learning.

\begin{figure}[htbp]
    \centerline{
        \includegraphics[width=88mm]{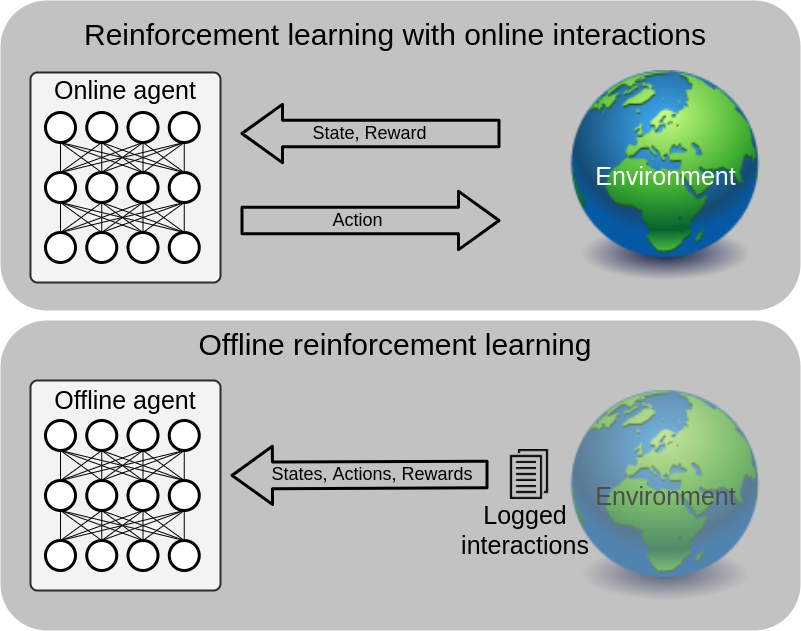}
        }
    \caption{Online and Offline RL interaction difference. An online agent can perform a selected action and observe its impact in the environment through state and reward signals. An offline agent passively receives a batch of logged interactions that consist of actions, states, and rewards signals.}
    \label{fig_online_offline_RL}
\end{figure}

Deployment efficiency measure \cite{Matsushima2020} in RL counts the number of changes in the data-collection policy during learning, i.e., an offline RL setup corresponds to a single deployment allowed for learning. Reinforcement learning methods can be classified through data and interaction perspectives \cite{lange2012batch}. What is named here as offline RL can be defined as pure batch algorithms, being classified as batch (data perspective) and offline (interaction perspective) methods. We refer to it as offline RL, since the use of a "batch" in an iterative learning algorithm can also refer to a method that consumes a batch of data, updates a model, and then obtains online a different batch, as opposed to a traditional online learning algorithm, which consumes one sample at a time \cite{Levine}. The interaction differences are shown in Fig.~\ref{fig_online_offline_RL}.

Given that only a fixed (finite) set of samples are available, note that the objective of offline RL is to learn the best possible policy from the given data and not an optimal policy, as in general reinforcement learning. Since there is no possibility of improving exploration, which is the main difficulty in offline RL. Samples distribution is another challenge that the offline RL approach needs to tackle. Generally, RL assumes that experiences samples are from a representative distribution of the environment, but the agent in offline RL has no control or knowledge of how they are sampled. While we cannot expect offline RL to discover actions that are better than any action in logged data, we can expect it to effectively utilize the compositional structure inherent in any temporal process \cite{Levine}.

Offline RL needs to learn a policy that does something differently, presumably better, from the behavior pattern observed in the experience samples \cite{Levine}. Therefore, changes in visited states and actions mean be evaluated on a distribution different from the training one, which forces offline RL to not assume that the data is independent and identically distributed (i.i.d.). Recently, fully offline reinforcement learning methods like Random Ensemble Mixture (REM) \cite{Agarwal2019}, Deep Q-learning from Demonstrations (DQfD) \cite{Hester2018}, Bootstrapping Error Accumulation Reduction (BEAR) \cite{Kumar2019}, Batch-Constrained deep Q-learning (BCQ) \cite{Fujimoto,Fujimoto2018}, and Behavior Regularized Actor Critic (BRAC) \cite{Wu2019} consider different approaches and techniques to surpass those limitations. Behavior-Regularized Model-ENsemble (BREMEN) \cite{Matsushima2020} is a model-based algorithm that can be used in a fully offline setup, but its main goal is to be efficient considering the number of changes in the data-collection policy needed during learning (deployment efficiency) and sampling efficient by using a mixed (online and offline) approach. Similar to BREMEN, Advantage Weighted Actor Critic (AWAC) \cite{Nair2020AcceleratingOR} is focused on effectively fine-tuning using online experiences after an offline pre-training period.

In deep reinforcement learning, for large or continuous state and action spaces we can represent the various components of agents, such as policies $\pi(s, a)$ or state-action values $Q(s, a)$, as an approximation with neural networks, e.g. Dueling Double Deep Q-Network (Dueling DDQN) \cite{wang2016dueling,hessel2018rainbow}. In Dueling DDQN, the classical DQN \cite{mnih2015human} is extended using Dueling Networks and Double Q-learning\cite{van2015ddqn} techniques.

Each policy $\pi$ has a corresponding state-action function $Q_{\pi} (s, a) = \mathbb{E}_{\pi}[R_t|s, a]$, the expected return when following the policy after taking action $a$ in state $s$, and a state value function $V_{\pi}(s) = \mathbb{E}_{a \sim \pi(s)}[Q_{\pi}(s,a)]$, that measures how good it is to be in a particular state $s$. In Double Q-Learning, an approximation of the deep Q-function $Q_{\pi}(s, a; \theta) = Q_{\pi}^{\theta}(s,a)$ with parameters $\theta$ is used. However, the value function $Q_{\pi}^{\theta}$ is updated using the target $Q^{\bar{\theta}}_{\pi}(s', a')$  as

\begin{equation}
    \label{form_y_ddqn}
y^{\text{DDQN}}_i = R_{t+1} + \gamma_{t+1} Q_{\pi}^{\bar{\theta}}(s', \underset{a'}{\mathrm{arg\,max\,}} Q_{\pi}^{\theta_i} (s', a')),
\end{equation}
where $\bar{\theta}$ represents the parameters of a fixed and separate target network. Moreover, $s'$ and $a'$ are state and action at $t+1$, respectively.

The parameters of the neural network are optimized by using stochastic gradient descent to minimize the loss 

\begin{equation}
    \label{form_loss_ddqn}
\mathcal{L}(\theta) = \|y^{\text{DDQN}}_i - Q_{\pi}^{\theta} (s, a)\|^ {2}.
\end{equation}

The gradient of the loss is back-propagated only into the parameters $\theta$ of the online network, which is also used to select actions. Periodically, the target network parameters $\bar{\theta}$, which is not directly optimized, are updated as $\bar{\theta} \gets \tau\theta + (1-\tau)\bar{\theta}$, where $\tau$ is the target update rate.

Dueling Networks represents the value function $V_{\pi}(s)$ and the advantage function $A_{\pi}(s, a) = Q_{\pi}(s) - V_{\pi}(s)$ as two separated streams with a single deep model whose output combines the two to produce a state-action value $Q_{\pi}(s, a)$ \cite{wang2016dueling}. The streams are constructed such that they have the capability of providing separate estimates of the value and advantage functions.

The learning algorithm is provided with a static dataset $\mathcal{D} = \{(s^i_t, a^i_t, s^i_{t+1}, r^i_t)\}_{i=1}^{m}$ as a set of $m$ transitions and must learn the best policy it can using this dataset. When training the Q-network, instead of only using the current experience as prescribed by standard temporal difference learning, the network is trained by sampling mini-batches of experiences from $\mathcal{D}$ uniformly at random. The sequence of losses thus takes the form
 
\begin{equation}
    \label{form_loss_offline_ddqn}
\mathcal{L}_i(\theta_i) = \mathbb{E}_{(s,a,r,s') \sim \mathcal{U}(\mathcal{D})} [(y^{\text{DDQN}}_i - Q_{\pi}^{\theta_i} (s, a; \theta_i))^ {2}].
\end{equation}

\subsection{Discrete state space}
As some algorithms are not suitable for continuous state space and simplification, the state space is discretized through clustering. K-means clustering algorithm \cite{arthur2006kmeans} can discretize the state space and the number of clusters can be defined by using the Bayesian information criterion (BIC) or Akaike information criterion (AIC) \cite{Komorowski2018}. 

Clustering with the K-means method, and its variants, can only produce convex clusters\cite{mitra2003nonconvex}. Alternatives can be applied as the OPTICS \cite{ankerst1999optics} method, which is a density-based clustering algorithm suitable for an arbitrary shape of cluster\cite{nagpal2013cluster_review}.

In this work, discrete state space is created using X-means\cite{Pelleg2015}, a K-means variant, with BIC, therefore there is no need to manually define a specific number of clusters. Moreover, another discrete state space is created using OPTICS for comparison.

\subsection{Off-policy policy evaluation (OPE)}
Offline reinforcement learning algorithms are trained with offline data, however, the agent must perform in real situations. Although, in some problems, evaluation can be performed through simulation, often the same restrictions that led to the use of offline RL apply in evaluation. So, it is important to predict how well the new policy will perform before deploying it.

Off-policy policy evaluation tackles the performance prediction problem producing an estimate that minimizes some concept of error \cite{Thomas}. Alternatives to OPE are, for example, crowd-sourced human labeling agent actions \cite{Jaques2019}, an expert qualitative analysis \cite{Raghu2017} and policy ranking \cite{Wu2019}.

Recently, supervised learning has been applied to large and diverse training datasets available \cite{goodfellow2016deep}. There are few initiatives to establish an offline RL benchmark as \cite{Fu2020, Wu2019, gulcehre2020rl}, but the field still lacks realistic evaluation protocols  \cite{Fu2020}.

Off-policy policy evaluation is an active field of research and multiples estimators can be applied. Two of them are Sequential Weighted Doubly-Robust (SWDR) and Model and Guided Importance Sampling Combining (MAGIC) \cite{Thomas} estimators to evaluate and compare offline RL methodologies. Both are Doubly-Robust (DR) based methods specifically designed for evaluating policies on RL problems where the horizon of episodes is longer than one. Besides the low bias advantage inherited from the DR method \cite{dudik2011dr}, if either action propensities or the reward function is accurate, those estimators balance the bias-variance trade-off while maintaining asymptotic consistency \cite{Thomas}.

For a description of MAGIC and SWDR methods for evaluation, let an approximate model of an MDP, $\widehat{r}^{\pi}(s,a,t)$ denotes the model's prediction of the reward $t$ steps later, $R_t$, if $S_0 = s$, $A_0 = a$, and the policy $\pi$ is used to generate the subsequent actions. Let
\begin{equation*}
    \label{form_estimate_r}
        \widehat{r}^{\pi}(s,t) = \displaystyle \sum_{a \in \mathcal{A}} \pi(a|s) \widehat{r}^{\pi}(s,a,t)
\end{equation*}
be a prediction of $R_t$ if $S_0 = s$, $A_0 = a$, and the policy $\pi$ is used to generate actions $A_0, A_1, \dots$, for all $(s,t,\pi) \in \mathcal{S} \times \mathbb{N}_{\geq 0} \times \Pi$. We can also define the estimated state value $\widehat{v}^{\pi}(s)$ and the estimate state-action value $\widehat{q}^{\pi}(s)$ as, respectively,

\begin{equation*}
    \label{form_estimate_v}
\widehat{v}^{\pi}(s) = \sum^{\infty}_{t=0} \gamma \widehat{r}^{\pi}(s,t),
\end{equation*}

\begin{equation*}
    \label{form_estimate_q}
\widehat{q}^{\pi}(s) = \sum^{\infty}_{t=0} \gamma \widehat{r}^{\pi}(s,a,t).
\end{equation*}

Given a historical dataset $D = \{H_i|\pi_i\}^n_{i=1}$ as a set of $n$ trajectories and known policies, called behavior policies, that generated them. A partial importance sampling estimator called off-policy $j$-step return\cite{Thomas}, $g^{(j)}(D)$, which uses an importance sampling based method to predict the outcome by using an evaluated policy $\pi_e$ up to $R_j$ is generated, and the approximate model estimator to predict the outcomes thereafter is proposed in\cite{Thomas}. That is, for all $j \in \mathbb{N}_{\geq -1}$ and using Weighted doubly-robust (WDR) method, let

\begin{equation}
    \label{form_WDR_g}
    \begin{split}
        g^{(j)}(D) = \displaystyle \sum^{n}_{i=1} \displaystyle \sum^{j}_{t=0} \gamma^t w^i_t R^{H_i}_t + \displaystyle \displaystyle \sum^{n}_{i=1} \gamma^{j+1} w^i_j \widehat{v}^{\pi_e} (S^{H_i}_{j+1}) \\ - \displaystyle \sum^{n}_{i=1} \displaystyle \sum^{j}_{t=0} \gamma^t (w^i_t \widehat{q}^{\pi_e} (S^{H_i}_t, A^{H_i}_t) - w^i_{t-1} \widehat{v}^{\pi_e}(S^{H_i}_t)),
    \end{split}
\end{equation}
where $n = |D|$, $S^H_t$
denotes the state at time $t$ during trajectory $H$ and $w^i_j$ is the weighted importance sampling

\begin{equation*}
    \label{form_WDR_w}
        w^i_j = \displaystyle \frac{\rho^i_t}{\displaystyle \sum^n_{j=1} \rho^j_t}
\end{equation*}
and $\rho_t(H, \pi_e, \pi_b) = \prod^{t}_{i=0} \frac{\pi_e(A^H_i|S^H_i)}{\pi_b(A^H_i|S^H_i)}$, is an importance weight, which is the probability of the first $t$ steps of $H$ under the evaluation policy $\pi_e$ divided by its probability under the behavior policy $\pi_b$. For brevity, $\rho_t(H_i, \pi_e, \pi_i)$ is written here as $\rho^i_t$. 

Notice that $g^{(-1)}(D)$ is a purely model-based estimator,
$g^{(\infty)}(D)$ is the WDR estimator, and the other off-policy $j$-step returns are partial WDR estimators that blend between these two extremes. So, the WDR estimator can be defined as

\begin{equation*}
    \label{form_WDR}
    \text{WDR}(D) = g^{(\infty)}(D) = \displaystyle \lim_{j \to \infty} g^{(j)}(D).
\end{equation*}

A Blending Importance Sampling and Model (BIM) estimator is defined as $BIM(D) = \mathbf{x}^{\top} g(D)$, where $\mathbf{x}= (x_{-1}, x_0, x_1, \dots)^{\top}$ is a weight vector and $g(D) = (g^{(-1)}(D), g^{(0)}(D), \dots)^{\top}$. So, we estimate $\mathbf{x}^*$ by minimizing an approximation of $MSE(\mathbf{x}^{\top} g(D), v(\pi_e))$. For the approximation, we use a subset of the returns, $\{g^{(j)}(D)\}$, for $j\in \mathcal{J}$, where $|\mathcal{J}| < \infty$. For all $j \notin \mathcal{J}$, we assign $\mathbf{x}_j = 0$. We also always include $-1$ and $\infty$ in $\mathcal{J}$. Let $\mathbf{g}_{\mathcal{J}}(D) \in \mathbb{R}^{|\mathcal{J}|}$ be the elements of $\mathbf{g}(D)$ whose indexes are in $\mathcal{J}$, the returns that will not necessarily be given weights of zero. Also let $\mathcal{J}_j$ denote the $j$\textsuperscript{th} element in $\mathcal{J}$. Before redefining the BIM estimator, we need to introduce the bias approximation $\widehat{\mathbf{b}}_n$ and the covariance approximation $\widehat{\Omega}_n$, when there are $n$ trajectories in $D$. 

After computing the percentile bootstrap $10\%$ confidence interval, $[l, u]$, for the mean of $g^{(\infty)}(D)$, which we ensure includes $\text{WDR}(D)$, the bias approximation is defined as

\begin{equation*}
    \label{form_WDR_bias}
    \widehat{\mathbf{b}}_n(j) \gets
                \begin{cases}
                    g^{(\mathcal{J}_i)}(D) - u,       & \quad \text{if } g^{(\mathcal{J}_i)}(D) > u \\
                    g^{(\mathcal{J}_i)}(D) - l,       & \quad \text{if } g^{(\mathcal{J}_i)}(D) < l \\
                    0,  & \quad \text{ otherwise.}
                \end{cases}
\end{equation*}

The covariance approximation $\widehat{\Omega}_n$ is defined as
\begin{equation}
    \label{form_WDR_omega}
        \begin{split}
            \widehat{\Omega}_n(i,j) = \displaystyle \frac{n}{n-1} \displaystyle \sum^n_{k=1} (g^{(\mathcal{J}_i)}_k(D) - \overline{g}^{(\mathcal{J}_i)}_k(D)) \\ \times (g^{(\mathcal{J}_j)}_k(D) - \overline{g}^{(\mathcal{J}_j)}_k(D)),
        \end{split}
\end{equation}
 where
 \begin{equation}
    \label{form_WDR_omega_where}
        \overline{g}^{(\mathcal{J}_i)}_k(D) = \displaystyle \frac{1}{n} \displaystyle \sum^n_{k=1} g^{(\mathcal{J}_i)}_k(D).
\end{equation}

That approximation can be summarized and redefine BIM estimator as

\begin{equation*}
    \label{form_approx_BIM}
    \text{BIM}(D, \widehat{\Omega}_n, \widehat{\mathbf{b}}_n) = (\widehat{\mathbf{x}}^{*})^{\top} \mathbf{g}_{\mathcal{J}}(D),
\end{equation*}
where
\begin{equation*}
    \label{form_approx_x_star}
    \widehat{\mathbf{x}}^{*} \in \underset{\mathbf{x} \in \Delta^{|\mathcal{J}|}}{\mathrm{arg\,min\,}} \mathbf{x}^{\top} [\widehat{\Omega_n} + \widehat{\mathbf{b}}_n \widehat{\mathbf{b}}^{\top}_n] \mathbf{x}.
\end{equation*}

Both estimators are designed to evaluate policies acting sequentially and are step-wise estimators. In SWDR a low bias can be achieved if either action propensities or the reward function is accurate. It uses weighted importance sampling and balances bias-variance trade-off while maintaining asymptotic consistency \cite{Gauci2018}. MAGIC also balances bias-variance trade-off using SWDR combined with a purely model-based estimator (blending importance sampling and model (BIM)) \cite{Thomas}.

\section{Methodology}
\label{sec:methodology}

This work uses the student evasion data from undergraduate students who started and completed the course between $2008$ and $2018$ at the Federal University of Espírito Santo (Ufes). The data consists of pseudo-anonymous academic, social, and demographic information presented in $13150$ observed semesters from a total of $1342$ students. Each observed semester is represented by the subjects taken in the period and has a total of $37$ features. Each student is represented by a sequence of academic terms. At the end of this sequence, the way of evasion is known and determined the student's outcomes, i.e, success or not.

An initial state space is defined as a $10$ continuous dimension state, consisting only of academics information and after a manual aggregation of all courses taken in each term, as listed in Table~\ref{tab_features_state}. After that, a discretization of the state features was used to represent the state space, where the discrete state space is the identifier of the cluster to which it belongs. So, it is a simplified one-dimension discrete state space defined through the clustering of those $10$ initial features. Two clustering algorithms are evaluated in this work by creating different discrete state spaces using X-means and OPTICS.

\begin{table}[htbp]
    \caption{State features}
    \begin{center}
      \centering
      \begin{tabular}{|p{36mm}|p{44mm}|}
        \hline
        Feature                              & Description  \\
        \hline
        CH CURSO                  & Total major’s course hours \\
        NUM PERIODOS SUGERIDO              & Major’s suggested terms \\
        NUM MAX PERIODOS               & Major’s maximum allowed terms \\
        MEDIA FINAL mean              & Mean of aggregated grades of all courses in the term \\
        MEDIA FINAL std              & Standard deviation of aggregated grades of all courses in the term \\
        CH DISCIPLINA mean                 & Mean of aggregated hours per course of all courses in the term \\
        CH DISCIPLINA std                 & Standard deviation of aggregated hours per course of all courses in the term \\
        NUM FALTAS mean   & Mean of aggregated  student’s  frequency  of  all  courses  in the term  \\
        NUM FALTAS std   & Standard deviation of aggregated  student’s  frequency  of  all  courses  in the term  \\
        COD DISCIPLINA count    & Number of completed courses \\
        \hline
      \end{tabular}

    \label{tab_features_state}
    \end{center}
\end{table}

\begin{figure}[htbp]
    \centerline{
        \includegraphics[width=88mm]{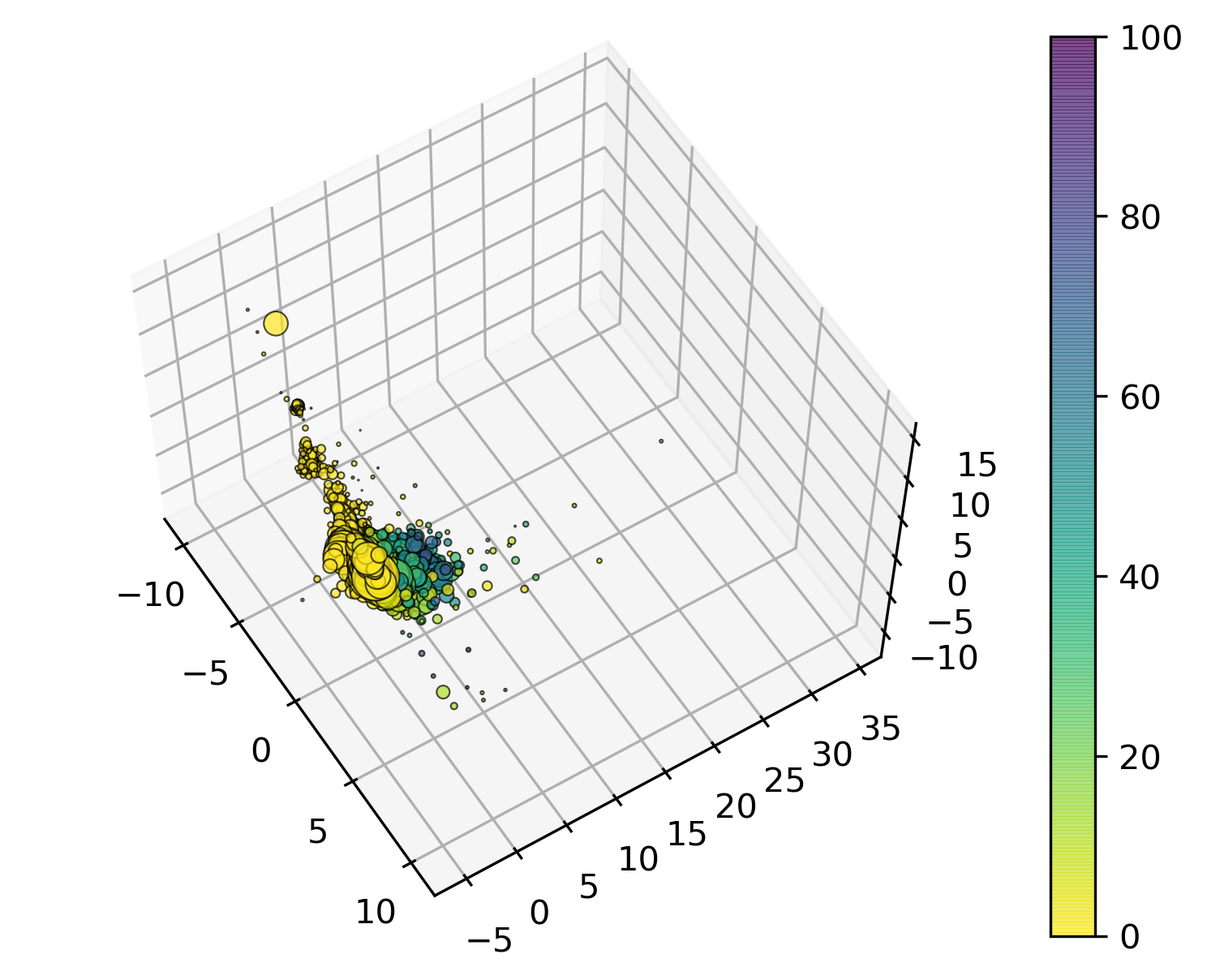}
        }
    \caption{State clusterization using X-means. Circle size represents state frequency in the dataset. Circle color represents the dropout rate in that state.}
    \label{fig_state_cluster_xmeans}
\end{figure}

The discretized states are represented in Fig.~\ref{fig_state_cluster_xmeans}, where the cluster centroids are expressed through a reduction of dimensionality to 3D using PCA. The frequency at which a given state occurs in the dataset is represented by the size of the circle, the higher the more frequent. The occurrence of trajectories that end in evasion is highlighted by the color of each circle, according to the scale in the percentile described in the sidebar of the figure. A spontaneous gradient in the dropout rate supports the validity of this state representation as it indicates that, in fact, there is an association between state membership and dropout rate. The OPTICS algorithm does not produce a clustering explicitly \cite{ankerst1999optics}, so, it does not define a center for each cluster and it is not possible to present a similar representation of its states.

A one-dimension discrete action state describes actions deployed to the student in that term. The action space consists of all possible combinations of two action features, which are the existence of a mandatory supervised study plan for that student ($2$ options) and the types of available student aid ($5$ options). In that manner, each action is represented by an index of all $10$ possibilities. In Ufes dataset only one action is possible per action type at each semester.

The reward function is a sparse function that returns non-zero signals only if the current state is the last state in the student trajectory and the student has a successful outcome. It is defined as:

\begin{equation}
    \label{form_SparsePReward}
    R(s_t,a_t) =
      \begin{cases}
        1,       & \text{if } s_t \text{ is a terminal academic}\\ & \quad \quad \text{successful state}\\
        0,        & \text{otherwise.}\\
      \end{cases}
\end{equation}

To create a new policy from the logged policy in the dataset, we used the Dueling Double DQN \cite{hessel2018rainbow} algorithm. This off-policy reinforcement learning algorithm can be setup to work online or offline. We deployed it in offline setup as needed for the aforementioned problem according to Algorithm~\ref{alg_Dueling_DDQN}. The performance of the newly learned policies is evaluated by off-policy policy evaluation of the reward. For that, we used MAGIC and SWDR  methods described by Algorithm~\ref{alg_MAGIC} and Algorithm~\ref{alg_SWDR}, respectively.

\begin{algorithm}
    \caption{Dueling  Double DQN\cite{hessel2018rainbow}}
    \label{alg_Dueling_DDQN}
    \begin{algorithmic}[1]
        \renewcommand{\algorithmicrequire}{\textbf{Input:}}
        \renewcommand{\algorithmicensure}{\textbf{Output:}}
        \REQUIRE~\\
        \begin{itemize}
            \item $\mathcal{D}$ - dataset of transitions;
            \item $\theta$ - initial network parameters, 
            \item $\tau$ - target update rate;
            \item $N_b$ - training batch size;
            \item $\bar{N}$ - target network replacement frequency.
        \end{itemize}
        
        \STATE $\bar{\theta} \gets \theta$
        \FOR {$t \in \{0, 1, \dots\}$}

            \STATE Sample a minibatch of $N_b$ tuples $(s, a, r, s') \sim \mathcal{U}(\mathcal{D})$.
            \STATE Construct target values, one for each of the $N_b$ tuples.
            \STATE $a^{max} (s'; {\bar{\theta}}) = \underset{a'}{\mathrm{arg\,max\,}} Q_{\pi}^{\bar{\theta}} (s', a')$
            
            \IF{$s'$ is terminal}
                \STATE $y_j = r$
            \ELSE
                \STATE $y_j = R_{t+1} + \gamma_{t+1} Q_{\pi}^{\theta}(s', a^{max} (s'; {\bar{\theta}}))$
            \ENDIF

            \STATE Do a gradient descent step according to Eq.~\ref{form_loss_offline_ddqn} loss every $\bar{N}$ steps.
            
            \STATE Update target parameters $\bar{\theta} \gets \tau\theta + (1-\tau)\bar{\theta}$ every $\bar{N}$ steps.
            \STATE Calculate OPEs according to Algorithm~\ref{alg_MAGIC} and Algorithm~\ref{alg_SWDR}.
        \ENDFOR
        
        \RETURN Newly learned policy $\pi$
    \end{algorithmic} 
\end{algorithm}

\begin{algorithm}
    \caption{Model and Guided Importance Sampling Combining (MAGIC) \cite{Thomas}}
    \label{alg_MAGIC}

    \begin{algorithmic}[1]
        \renewcommand{\algorithmicrequire}{\textbf{Input:}}
        \renewcommand{\algorithmicensure}{\textbf{Output:}}
        \REQUIRE~\\
        \begin{itemize}
            \item $D$: Historical data;
            \item $pi_e$: Evaluation policy;
            \item Approximate model that allows for computation of $\hat{r}^{\pi_e}(s,a,t)$;
            \item $\mathcal{J}$: The set of return lengths to consider;
            \item $k$: The number of bootstrap resampling.
            
        \end{itemize}
        
        \STATE Calculate $\widehat{\Omega}_n$ according to Eq.~\ref{form_WDR_omega}.
        \STATE Allocate $D_{(.)}$ so that for all $i \in \{1, \dots, k\}, D_i$ can hold $n$ trajectories. 
        
        \FOR {$i \in \{1, \dots k\}$}
            \STATE Load $D_i$ with $n$ uniform random samples drawn from $D$ with replacement.
        \ENDFOR
        
        \STATE $\mathbf{v} = sort(g^{(\infty)}(D_{(.)})$
        \STATE $l \gets \min\{\text{WDR}(D), \mathbf{v}(\lfloor 0.05n \rfloor) \}$
        \STATE $u \gets \max\{\text{WDR}(D), \mathbf{v}(\lceil 0.5n \rceil) \}$
        
        \FOR {$j \in \{1, \dots |\mathcal{J}|\}$}
            \STATE $\widehat{\mathbf{b}}_n(j) \gets
                \begin{cases}
                    g^{(\mathcal{J}_i)}(D) - u,       & \quad \text{if } g^{(\mathcal{J}_i)}(D) > u \\
                    g^{(\mathcal{J}_i)}(D) - l       & \quad \text{if } g^{(\mathcal{J}_i)}(D) < l \\
                    0,  & \quad \text{ otherwise.}
                \end{cases}$
        \ENDFOR
        
        \STATE $\mathbf{x} \gets \underset{\mathbf{x} \in \Delta^{|\mathcal{J}|}}{\mathrm{arg\,min\,}} \mathbf{x}^{\top} [\widehat{\Omega_n} + \widehat{\mathbf{b}}_n \widehat{\mathbf{b}}^{\top}_n] \mathbf{x}$
        \RETURN $\mathbf{x}^{\top}\mathbf{g}_{\mathcal{J}}(D)$
    \end{algorithmic} 
\end{algorithm}

\begin{algorithm}
    \caption{Sequential weighted doubly-robust (SWDR) \cite{Thomas}}
    \label{alg_SWDR}
    \begin{algorithmic}[1]
        \renewcommand{\algorithmicrequire}{\textbf{Input:}}
        \renewcommand{\algorithmicensure}{\textbf{Output:}}
        \REQUIRE~\\
        \begin{itemize}
            \item $D$: Historical data;
            \item $\pi_e$: Evaluation policy;
            \item Approximate model that allows for computation of $\hat{r}^{\pi_e}(s,a,t)$;
            \item $k$: The number of bootstrap resampling.
        \end{itemize}

    \STATE $\mathcal{J} = \{\infty\}$
    
    \RETURN $\text{MAGIC}(D, \pi_e, \hat{r}^{\pi_e}(s,a,t), \mathcal{J}, k)$

    \end{algorithmic} 
\end{algorithm}

\section{Experimental Results}
\label{sec:experimental_results}

In our tests, for each clustering method, we defined $3$ different discrete datasets. Then, we performed one trial for each discrete dataset. We also have performed a limited tuning starting with the values used in the paper \cite{hessel2018rainbow}. The offline setup of the Dueling Double DQN \cite{hessel2018rainbow} algorithm was used to create a new policy from the logged policy. The Q-network is a fully-connected layer of size $128$ followed by two parallel fully-connected layers of size $32$, creating a dueling architecture. The network is trained for $25$ epochs with minibatch size $512$. The optimizer used was ADAM with a learning rate of $0.01$ and a learning rate decay of $0.999$. The target update rate is $0.1$ and the discount factor is $0.99$. All experiments run in a computer with the following specifications: i5-8265U processor, 8 GB DDR4 RAM system memory, $1$ TB HD (SATA 3.1 $5400$ RPM) and $240$ GB SSD (PCIe NVMe Ger 3.0 x2) storage. Each discretization took $127.945$ seconds on average and each trial policy training took $38.117$ seconds on average.

\begin{figure}[!t]
    \centering
    \subfloat[]{\includegraphics[width=43mm]{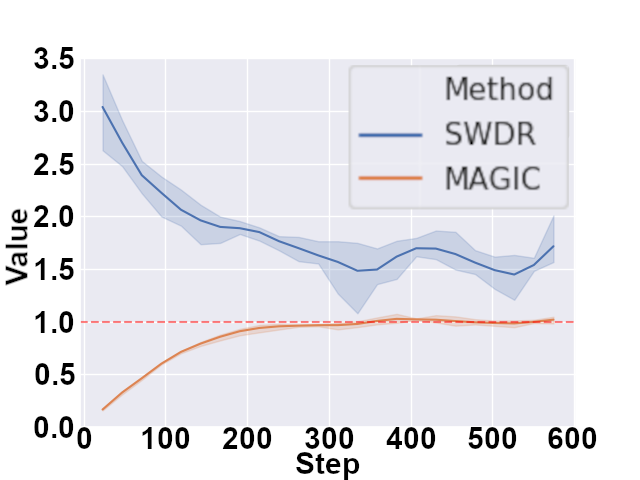}
    \label{fig_CPE_reward_xmeans}}
    \hfil
    \subfloat[]{\includegraphics[width=43mm]{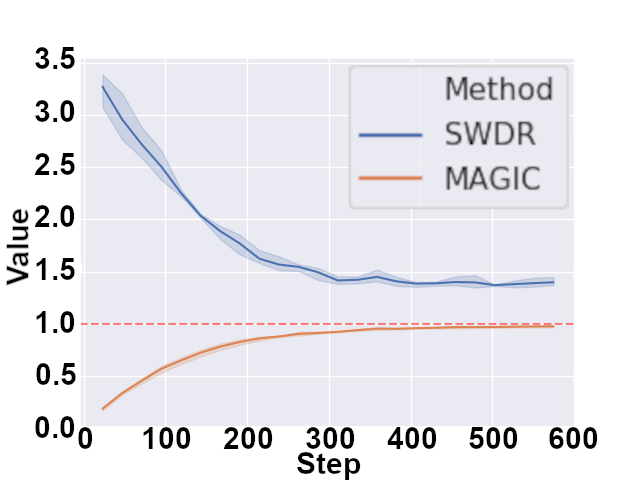}
    \label{fig_CPE_reward_optics}}
    \caption{OPE cumulative reward score (value axis), mean (solid line) and 95\% confidence (translucent error bands) over all trials, using X-means \protect\subref{fig_CPE_reward_xmeans} and OPTICS \protect\subref{fig_CPE_reward_optics} to discretize the state space. Notice that the RL model should achieve roughly $1.0$ to $1.5$ times as much cumulative reward as the logged policy in both scenarios.}
    \label{fig_CPE_reward}
\end{figure}

In Fig.~\ref{fig_CPE_reward_xmeans} and Fig.~\ref{fig_CPE_reward_optics} there is the OPE cumulative reward score of the policies using X-means and OPTICS, respectively, to discretize state space. Both show a solid line as the average of the performances along with the training steps and the translucent error bands as the $95\%$ confidence range. The OPE cumulative reward score (value axis) on the graph represents how many times the performance of the new policy represents the performance of the logged policy. Therefore, the dashed line ($value=1$) represents the point where the learned policy is equivalent to the logged policy, values above it mean that the new policy performs better and values below represent worse performance. 

In both figures, the SWDR estimator shows a value close to $1.5$, which is $50\%$ of improvement over logged policy, with the OPTICS method having a narrower $95\%$ confidence. However, the MAGIC estimator shows that the learned policies performed approximately equivalent to the logged policy. It also means that there was possible to learn a policy using both clustering methods to discretize the state space. Despite that, there is no guarantee of the effectiveness of the learned policies due to overestimation. There is the possibility of the occurrence of distributional shift, inducing overestimation if the learned policy does not stay close to the behavior policy \cite{paine2020hyperparameter}. Difficulties in properly estimate the policy performance before deploying it is a major issue in applying offline RL in real-world scenarios\cite{Levine}. That can explain the expressive difference between estimators.

\begin{figure}[!t]
    \centering
    \subfloat[]{\includegraphics[width=43mm]{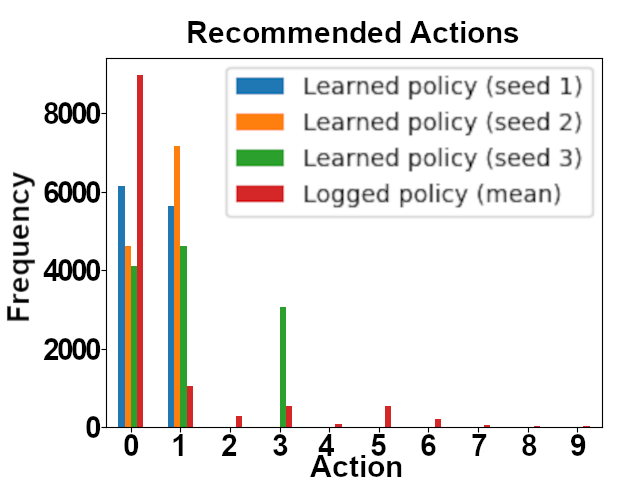}
    \label{fig_actions_bar_xmeans}}
    \hfil
    \subfloat[]{\includegraphics[width=43mm]{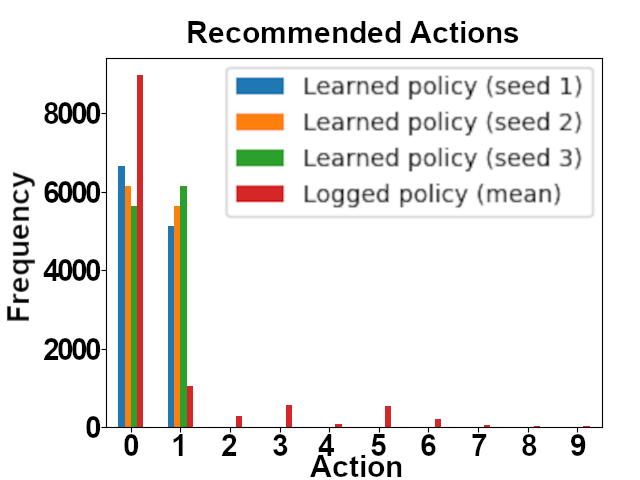}
    \label{fig_actions_bar_optics}}
    \caption{Count of occurrences of each possible action in evaluation set after training using X-means \protect\subref{fig_actions_bar_xmeans} and OPTICS \protect\subref{fig_actions_bar_optics} clustering. Logged policy actions frequency is expressed as the mean of the logged actions occurrence evaluated in all seeds.}
    \label{fig_actions_bar}
\end{figure}

A comparison between the actions taken by the new policies and the logged policy is shown in Fig.~\ref{fig_actions_bar_xmeans} and Fig.~\ref{fig_actions_bar_optics}, using X-means and OPTICS for clustering, respectively. Logged policy action frequencies are expressed as the mean of the logged actions occurrence evaluated in all seeds. Note that all learned policies focus on using almost only two actions, which may indicate that other actions do not have an impact on improving the reward or, due to their low occurrence (low exploration), it was not possible to learn their proper use.

\section{Conclusion}
\label{sec:conclusion}

Recent developments in the machine learning field provide new ways to address problems that are still present, such as student dropout. In this paper, a methodology is proposed to reduce student dropout through the use of a decision support method to select appropriate aid actions using offline reinforcement learning.

Our analyzes were performed using real data from students at a Brazilian university and show promising results in terms of the ability to produce an AI policy, at least equivalent to the logged one, as it should achieve roughly $1.0$ to $1.5$ times as much cumulative reward as the logged policy. However, due to overestimation in offline RL when the learned policy loses similarity to the policy that generated the dataset, there is no guarantee of the learned policy effectiveness. Selecting algorithms that encourage policies to stay close to the behavior policy can reduce that problem.

The proposed discretization of the state space seems suitable to learn a policy in offline RL context. This work also makes a comparison of the impact of two different clustering methods for that discretization on the result, showing similar performance.

For future work, an investigation to identify features that are the most informative using PCA analysis and which features are most suitable for the state space and the possibility of a continuous state space approach might improve performance. Providing interpretability for agent's decisions through explainable reinforcement learning and understanding their impacts seems promising. A further investigation in the cases in which students were able to avoid dropouts and clarify what actions and spans are appropriate can be insightful for decision-makers.

\bibliographystyle{IEEEtran}
\bibliography{IEEEabrv,paper}

\end{document}